\documentclass{article}

\usepackage{microtype}
\usepackage{graphicx}
\usepackage{subfigure}
\usepackage{booktabs} 
\usepackage{siunitx}

\usepackage{amsmath}
\usepackage{amssymb}
\usepackage{dsfont}

\usepackage{natbib}

\usepackage{hyperref}
\usepackage[capitalise]{cleveref}

\usepackage[accepted]{icml2018}

\newcommand{\ignore}[1]{}

\renewcommand{\vec}[1]{\mathbf{#1}}
\newcommand{\mat}[1]{\mathbf{#1}}
\newcommand{\tr}{^{\mkern-1.5mu\mathsf{T}}}

\newenvironment{bsmallmatrix}{\left[\begin{smallmatrix}}{\end{smallmatrix}\right]}

\newcommand{\Dataset}{\mathcal{D}}
\newcommand{\Reals}{\mathds{R}}
\newcommand{\SimMetric}{\rho}  
\newcommand{\Fcn}{f}
\newcommand{\FeatFcn}{z}
\newcommand{\FeatParam}{\theta_{z}}
\newcommand{\FeatSpace}{\mathcal{Z}}
\newcommand{\ClfFcn}{g}
\newcommand{\ClfParam}{\theta_g}

\newcommand{\Loss}[1]{\mathcal{L}_{\mathrm{#1}}}

\newcommand{\bphi}{\boldsymbol{\phi}}
\newcommand{\bPhi}{\boldsymbol{\Phi}}

\begin{document}

\twocolumn[
\icmltitle{Semi-Supervised Learning via Compact Latent Space Clustering}



\icmlsetsymbol{equal}{*}

\begin{icmlauthorlist}
\icmlauthor{Konstantinos Kamnitsas}{msr,imp}
\icmlauthor{Daniel C. Castro}{msr,imp}
\icmlauthor{Loic Le Folgoc}{imp}
\icmlauthor{Ian Walker}{imp}
\icmlauthor{Ryutaro Tanno}{msr,ucl}
\icmlauthor{Daniel Rueckert}{imp}
\icmlauthor{Ben Glocker}{imp}
\icmlauthor{Antonio Criminisi}{msr}
\icmlauthor{Aditya Nori}{msr}
\end{icmlauthorlist}

\icmlaffiliation{msr}{Microsoft Research Cambridge, United Kingdom}
\icmlaffiliation{imp}{Imperial College London, United Kingdom}
\icmlaffiliation{ucl}{University College London, United Kingdom}

\icmlcorrespondingauthor{Konstantinos Kamnitsas}{konstantinos.kamnitsas12@imperial.ac.uk}

\icmlkeywords{Machine Learning, ICML}

\vskip 0.3in
]



\printAffiliationsAndNotice{}  


\begin{abstract}

We present a novel cost function for semi-supervised learning of neural networks that encourages compact clustering of the latent space to facilitate separation. The key idea is to dynamically create a graph over embeddings of labeled and unlabeled samples of a training batch to capture underlying structure in feature space, and use label propagation to estimate its high and low density regions.
We then devise a cost function based on Markov chains on the graph that regularizes the latent space to form a single compact cluster per class, while avoiding to disturb existing clusters during optimization.
We evaluate our approach on three benchmarks and compare to state-of-the art with promising results. Our approach combines the benefits of graph-based regularization  with efficient, inductive inference, does not require modifications to a network architecture, and can thus be easily applied to existing networks to enable an effective use of unlabeled data.

\end{abstract}


\section{Introduction}
\label{introduction}

\begin{figure}[!ht]
\centering
\includegraphics[width=\linewidth]{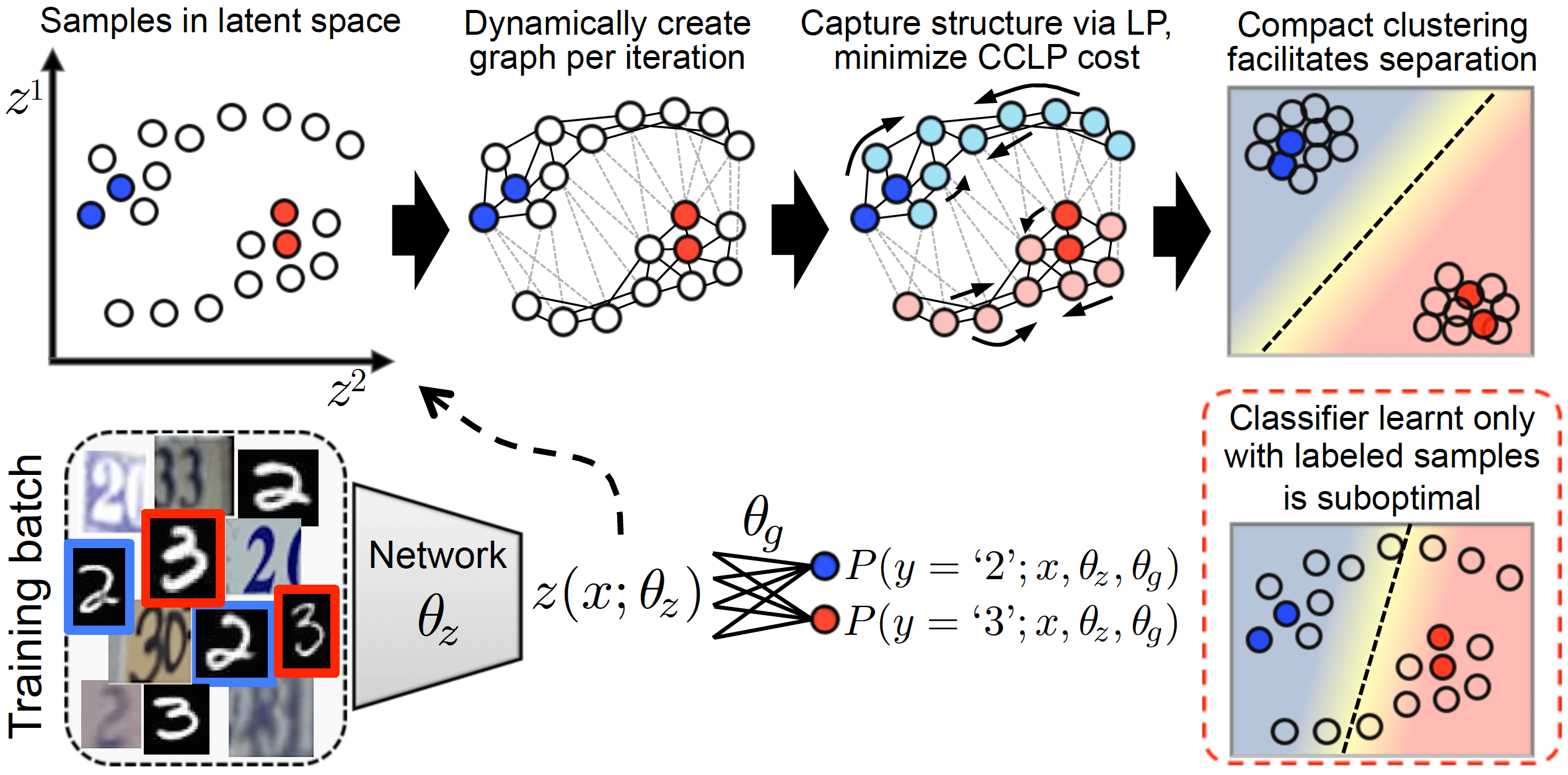}
\caption{Overview of our method. We dynamically construct a graph in the latent space of a network at each training iteration, propagate labels to capture the manifold's structure, and regularize it to form a single, compact cluster per class to facilitate separation.}
\label{fig:main}
\end{figure}

Semi-supervised learning (SSL) addresses the problem of learning a model by effectively leveraging both labeled and unlabeled data \cite{chapelle2009semi}. SSL is effective when it results in a model that generalizes better than a model learned from labeled data only. More formally, let $\mathcal{X}$ be a sample space with data points and $\mathcal{Y}$ the set of labels (e.g., referring to different classes). Let $\Dataset_L \subseteq \mathcal{X} \times \mathcal{Y}$ be a set of labeled data points, and let $\Dataset_U \subseteq \mathcal{X}$ be a set of unlabeled data. In this work we focus on classification tasks, where for each $(\vec{x},y) \in \Dataset_L$, $y$ is the ground truth label for sample $\vec{x}$.
Our objective is to learn a predictive model $f(\vec{x};\theta) = p(y|\vec{x},\theta)$, parametrized by $\theta$, which approximates the true conditional distribution $q(y|\vec{x})$ generating the target labels.
SSL methods learn this by utilizing both $\Dataset_L$ and $\Dataset_U$, often assuming that $|\Dataset_U| \gg |\Dataset_L|$. Thus leveraging the ample unlabeled data allows capturing more faithfully the structure of data.
 
Various approaches to SSL have been proposed (see \cref{sec:related} for an overview). Underlying most of them is the notion of \emph{consistency} \cite{zhou2004learning}: samples that are close in feature space should be close in output space (\emph{local consistency}) and samples forming an underlying structure should also map to similar labels (\emph{global consistency}). This is the essence of the \textit{smoothness} and \textit{cluster assumptions} in SSL \cite{chapelle2009semi} that underpin our work. They respectively state that the label function should be smooth in high density areas of feature space, and points that belong to the same cluster should be of the same class. Hence decision boundaries should lie in low density areas.

We present a simple and effective SSL method for regularizing inductive neural networks (\cref{fig:main}). The main idea is to dynamically create a graph in the network's latent space over samples in each training batch (containing both labeled and unlabeled data) to model the data manifold as it evolves during training.
We then regularize the manifold's structure globally towards a more favorable state for class separation.
We argue that the optimal feature space for classification should cluster all examples of a class to a single, compact component, proposing a further constraint for SSL to those previously discussed: all samples that map to the same class should belong to a single cluster.
To learn such a latent space, we first use label propagation (LP) \cite{zhu2002learning} as a proxy mechanism to estimate the arrangement of high/low density regions in latent space. This is in contrast to using LP as a transductive inference mechanism as done previously.
We then propose a novel cost function, formulated via Markov chains on the graph, which not only brings together parts of the manifold with similar estimated LP posterior to form compact clusters, but also defines an optimization process that avoids disturbing existing high density areas, which are manifestations of information important for SSL.

We evaluate our approach on three visual recognition benchmarks: MNIST, SVHN and CIFAR10. Despite its simplicity, our method compares favorably to current state-of-the-art SSL approaches when labeled data is limited. Moreover, our regularization offers consistent improvements over standard supervision even when the whole labeled set is used. Our technique is computationally efficient and does not require additional network components. Thus it can be easily applied to existing models to leverage unlabeled data or regularize fully supervised systems.


\section{Related Work}
\label{sec:related}

The great potential and practical implications of utilizing unlabeled data has resulted in a large body of research on SSL. The techniques can be broadly categorized as follows.

\subsection{Graph-Based Methods}
\label{subsec:related_graph}

These methods operate over an input graph with adjacency matrix $\mat{A}$, where element $A_{ij}$ is the similarity between samples $\vec{x}_i, \vec{x}_j \in {\Dataset_L \cup \Dataset_U}$. Similarity can be based on Euclidean distance \cite{zhu2002learning} or other, sometimes task-specific metrics \cite{weston2012deep}. Transductive inference for the graph's unlabeled nodes is done based on the smoothness assumption, that nearby samples should have similar class posteriors. Label propagation (LP) \cite{zhu2002learning} iteratively propagates the class posterior of each node to neighbors, faster through high density regions, until a global equilibrium is reached. \citet{zhu2003semi} showed that for binary classification one arrives at the same solution by minimizing the energy:
\begin{equation}\label{eq:energy_lp}
E(\vec{f}) = \frac{1}{2} \sum_{i,j} A_{ij} ( f(\vec{x}_i) - f(\vec{x}_j) )^2 = \vec{f}^{\top} \Delta \vec{f} \,.
\end{equation}
Here, $\vec{f}$ is the vector with responses from predictor $f(\vec{x})=p(y=0|\vec{x})$ applied to all samples, $\Delta$ is the graph Laplacian. The solution being a harmonic function implies that the resulting posteriors for unlabeled nodes are the average of their neighbors \cite{zhu2005semi}, showing that LP agrees with the smoothness assumption. \citet{zhou2004learning} proposed a similar propagation rule and argued that predictions by propagation agree with the notion of global consistency. Many variations followed, such as the diffusion and graph convolutional networks \cite{atwood2016diffusion,kipf2016semi}. These approaches are \emph{transductive} and require a \emph{pre-constructed} graph as a given, while their performance largely relies on the suitability of this given graph for the task. In contrast, we use LP not for transductive inference but as a sub-routine to estimate the structure of the clusters in a network's latent space. We then regularize the network's feature extractor, which \emph{learns} an appropriate graph consistent with the smoothness property of LP, while preserving the network's efficient \emph{inductive} classifier.

\Cref{eq:energy_lp} has also been used to define the graph Laplacian regularizer, which has been used for SSL of inductive models \cite{belkin2006manifold,weston2012deep}. However, these methods still require a pre-constructed graph. A recent method inspiring our work avoids this requirement and seeks associations between labeled and unlabeled data~\cite{haeusser2017learning}. This is modeled as a two-step random walk in feature space that starts and ends at labeled samples of the same class, via one intermediate unlabeled point. The method was not formulated via graphs but is related, as it models pairwise relations. But its formulation does not capture the global structure of the data, unlike ours, and can collapse to the trivial solution of associating an unlabeled point to its closest cluster in Euclidean space \cite{haeusser2017learning}. Hence, a second regularizer is required to keep all samples relatively close.

\subsection{Self-Supervision and Entropy Minimization}
\label{subsec:related_selfsup}

One of the earliest ideas for leveraging unlabeled data is self-supervision or self-learning. It is a wrapper framework in which a classifier trained with supervision periodically classifies the unlabeled data, and confidently classified samples are added to the training set. The idea dates back to \citet{scudder1965probability} and saw multiple extensions. The method is heavily dependent on classifier's performance. It gained popularity recently for training neural networks \cite{lee2013pseudo}, enabled by their overall good performance. Relevant is co-training \cite{blum1998combining}, which uses confident predictions of two classifiers trained on distinct views of the data.

Closely related is regularization via conditional entropy minimization \cite{grandvalet2005semi}. Model parameters $\theta$ are learned by minimizing entropy in the prediction $H(y|\vec{x},\theta) = \mathbb{E} \left[ -\log p(y|\vec{x},\theta) \right] $ for each unlabeled sample $\vec{x}$, additionally to the supervised loss. It can be seen as an efficient information-theoretic form of self-supervision, encouraging the model to make confident predictions. This pushes samples away from decision boundaries and vice-versa, favoring low-density separation. It may however induce confirmation bias, hurting optimization if clusters are not yet well formed. Such is the case of a neural network's embedding in early training stages, where gradient descent can push samples away from the decision boundary towards the random side where they started (\cref{fig:two_circles}). Because of this, the regularizer's effect is commonly controlled with ad-hoc ramp-up schedules of a weight meta-parameter \cite{springenberg2015, chongxuan2017triple, dai2017good, miyato2017virtual}. Similar is the case of self-supervision.
In contrast, our regularizer does not use the suboptimal classifier being trained. It only reasons about the latent manifold's geometry. As a result, gradients it applies are indifferent to the decision boundary's position and do not generally oppose gradients of the classification loss.
Finally, since our cost depends on the confidence of labels propagated on the graph, its effect adapts throughout training, according to whether clusters are well formed or not.

\subsection{Perturbation-Based Approaches}

Regularizing the input-output mapping to be consistent when noise is applied to the input can improve generalization \cite{bishop1995training}. This goal of ``consistency under perturbation'' has been shown applicable for SSL \cite{bachman2014learning}. In its generic form, a function $f$ minimizes a regularizer of the form $R(f) = \mathbb{E}_{\xi} \left[ d( f(\vec{x};\xi), f(\vec{x}) ) \right]$ for each sample $\vec{x}$, assuming $\xi$ is a noise process such that $ \mathbb{E}_{\xi} \left[f(\vec{x};\xi)\right] = f(\vec{x})$. $f$ can be the classification output or hidden activations of a neural network, $d$ is a distance metric such as the $L_2$ norm. This cost encourages \emph{local} consistency of the classifier's output around each unlabeled sample, pushing decision boundaries away from high density areas. The approach has given promising results, with $\xi$ taking various forms such as different dropout masks \cite{bachman2014learning}, Gaussian noise applied to network activations \cite{rasmus2015semi}, sampling input augmentations, predictions from models at different stages of training \cite{laine2016temporal,tarvainen2017mean} or adversarial perturbation \cite{miyato2017virtual}. Like self-supervision, these methods can induce confirmation bias. Orthogonally to encouraging local smoothness around each individual sample, our method regularizes geometry of the manifold globally by treating all samples and their connections jointly.

\subsection{Generative Models}

Generative models have also been used within SSL frameworks. In particular, probabilistic models such as Gaussian mixtures \cite{mclachlan2004discriminant} are representative examples. These approaches model how samples $\vec{x}$ are generated, estimating $p(\vec{x}|y)$ or the joint distribution $p(\vec{x},y)=p(\vec{x}|y)p(y)$. In this framework, SSL can be modeled as a missing data ($y$) problem. This is however a substantially more general problem than estimating $p(y|\vec{x})$ with a discriminative model. One might argue that estimating the joint distribution is not the best objective for SSL, as it requires models of unnecessarily large representational power and complexity. Examples of popular neural models are auto-encoders (AE) \cite{ranzato2008semi,rasmus2015semi} and variational auto-encoders (VAE) \cite{kingma2014semi,maaloe2016auxiliary}. Unfortunately, spending the encoder's capacity on preserving variation of the input that is potentially unrelated to label $y$, as well as the requirement for a similarly powerful decoder, make these approaches difficult to scale to large and complex databases.

Generative adversarial networks (GAN) \cite{goodfellow2014generative} have been recently applied to SSL with promising results. Conditional GANs were used to generate synthetic samples $(\vec{x},y)$, which can serve as additional training data \cite{chongxuan2017triple}. \citet{salimans2016improved} encouraged the discriminator to identify the class of real samples, aside from distinguishing real from fake inputs. Similarly, in CatGAN \cite{springenberg2015} the discriminator minimizes the conditional entropy of $p(y|\vec{x})$ for real but maximizes it for fake samples. The reason why the classification objective gains from the real-versus-fake discrimination was analyzed in \citet{dai2017good}. Interestingly, rather than directly benefiting from modeling the generative process, it was shown that bad examples from the generator that lie in low-density areas of the data manifold guide the classifier to better position its decision boundary, thus connecting the improvements with the cluster assumption. Promising results were achieved, yet the requirement for a generator and the challenges of adversarial optimization leave space for future work. Note that these methods are orthogonal to ours, which regularizes the latent manifold's structure.


\section{Method}
\label{sec:method}

Our work builds on the \emph{cluster assumption}, whereby samples forming a structure are likely of the same class \cite{chapelle2009semi}, by enforcing a further constraint:\\
\emph{All samples of a class should belong to the same structure}.

In this work we take the labeling function $\Fcn(\vec{x}; \theta)$ to be a multi-layer neural network. This model can be decomposed into a \emph{feature extractor} $\FeatFcn(\vec{x}; \FeatParam) \in \mathcal{Z}$ parametrized by $\FeatParam$, and a \emph{classifier} $\ClfFcn( \FeatFcn(\vec{x}; \FeatParam); \ClfParam)$ with parameters $\ClfParam$. The former typically consists of all hidden layers of the network, while the latter is the final linear classifier.
We argue that classification is improved whenever data from each class form compact, well separated clusters in \emph{feature space} $\FeatSpace$.
We use a graph embedding to capture the structure of data in this latent space and propagate labels to unlabeled samples through high density areas (\cref{subsec:capturing_shape}). We then introduce a regularizer (\cref{subsec:loss_func}) that 1) encourages compact clustering according to propagated labels and 2) avoids disturbing existing clusters during optimization (\cref{fig:main}).

\subsection{Estimating Structure of Data via Dynamic Graph
Construction and Label Propagation}
\label{subsec:capturing_shape}

We train $\Fcn$ with stochastic gradient descent (SGD), sampling at each SGD iteration a labeled batch $(\mat{X}_L, \vec{y}_L) \sim \Dataset_L $ of size $N_L$ and an unlabeled batch $\mat{X}_U \sim \Dataset_U $ of size $N_U$. Let $\mat{Y}_L \in \Reals^{N_L \times C} $ be one-hot representation of $\vec{y}_L$ with $C$ classes. The feature extractor of the network produces the embeddings $\mat{Z}_L = \FeatFcn(\mat{X}_L; \FeatParam)$ for labeled and $\mat{Z}_U = \FeatFcn(\mat{X}_U; \FeatParam)$ for unlabeled data. We propose to dynamically create a graph at every SGD iteration over the embedding $\mat{Z}= \begin{bsmallmatrix} \mat{Z}_L \\ \mat{Z}_U \end{bsmallmatrix}$ of batch $\mat{X}= \begin{bsmallmatrix} \mat{X}_L \\ \mat{X}_U \end{bsmallmatrix}$, and use label propagation (LP) \cite{zhu2002learning} to implicitly capture the structure of each class. Unlike Euclidean metrics, graph-based metrics naturally respect the underlying data distribution, following paths along high density areas.

We first generate a fully connected graph in feature space from both labeled and unlabeled samples. The graph is characterized by the adjacency matrix $\mat{A} \in \Reals^{N \times N}$, where $N=N_L+N_U$. Each element $A_{ij}$ is the weight of an edge between samples $i$ and $j$ representing their similarity, and is parametrized as
\begin{equation} \label{eq:adjacency_mat}
A_{ij} = \exp ( {\SimMetric(\vec{z}_i, \vec{z}_j)} ) \,, \quad
\forall \vec{z}_i, \vec{z}_j \in \mat{Z}_L \cup \mat{Z}_U \,,
\end{equation}
where $\SimMetric: \mathcal{Z}^2 \to \Reals$ is a similarity score such as the dot product or negative Euclidean distance. In this paper we use the former. The Markovian random walk along the nodes of this graph is defined by its transition matrix $\mat{H}$, obtained by row-wise normalization\footnote{We note that other LP variants such as \citet{zhou2004learning} that uses symmetrically normalized Laplacian could also be used.} of $\mat{A}$. Each element $H_{ij}$ is the probability of a transition from node $i$ to node $j$:
\begin{equation} \label{eq:transition_mat}
H_{ij} = A_{ij} \Big/ \sum_{k} A_{ik} \,.
\end{equation}
Without loss of generality, elements of $\mat{A}$ and $\mat{H}$ that correspond to labeled samples $L$ and unlabeled samples $U$ are arranged so that
\begin{equation}
\mat{H} = 
\begin{bmatrix}
\mat{H}_{LL} & \mat{H}_{UL} \\
\mat{H}_{LU} & \mat{H}_{UU}
\end{bmatrix} .
\end{equation}

LP uses $\mat{H}$ to model the process of a node $i$ propagating its class posterior $\bphi_{i} = p_{\mathrm{LP}}(y|\vec{x}_i,\mat{A}) \in \Reals^{C}$ to the other nodes. One such propagation step is formally given by $\bPhi^{(t+1)} = \mat{H} \bPhi^{(t)}$, where $\bPhi^{(t)} \in \Reals^{N \times C}$. As a result, class confidence propagates from labeled to unlabeled samples. While propagation to nearby points initially dominates due to the exponential in \cref{eq:adjacency_mat}, multiple iterations of the algorithm allow soft labels $\bPhi^{(t)}$ to propagate and eventually traverse the whole graph in the stationary state. Unlike \emph{diffusion} \cite{kondor2002diffusion}, LP interprets labeled samples as \emph{constant sources} of labels, and clamps their confidence to their true value $\mat{Y}_L$, thus $\bPhi^{(t)} = \begin{bsmallmatrix} \mat{Y}_L \\ \bPhi_U^{(t)} \end{bsmallmatrix}, \forall t$. Hence class confidence gradually accumulates in the graph. By propagating more easily through high density areas, the process converges at an equilibrium where the decision boundary settles in a low-density area, satisfying the cluster assumption. Conveniently, class posteriors for the unlabeled data at equilibrium can be computed in closed form \cite{zhu2002learning}, without iterations, as
\begin{equation} \label{eq:lp_phu_u}
\bPhi_U = (\mat{I} - \mat{H}_{UU})^{-1} \mat{H}_{UL} \mat{Y}_{L} \,.
\end{equation}

Hereafter, let $\bPhi = \begin{bsmallmatrix} \mat{Y}_L \\ \bPhi_U \end{bsmallmatrix} \in \Reals^{N \times C}$ denote the class posteriors estimated by LP at convergence, i.e.\ the concatenation of the true, hard (clamped) posteriors for $\mat{X}_L$ and the estimated posteriors for $\mat{X}_U$. \Cref{eq:lp_phu_u} has been previously used for \emph{transductive} inference in applications where the graph is given \emph{a priori} (see \cref{subsec:related_graph}), hence results directly rely on \emph{suitability} of the graph and LP for predictions. In contrast, we build the graph in feature space $\FeatSpace$ that will be \emph{learned} appropriately. We here point out that equation (\ref{eq:lp_phu_u}) is \emph{differentiable}. This enables learning $\FeatSpace$ that simultaneously complies with properties of LP while serving the optimization objectives. We also emphasize that instead of relying on it for inference, in our framework LP merely provides a mechanism for capturing the arrangement of clusters in latent space to regularize them towards a desired stationary point. This improved embedding will benefit generalization of the actual classifier $\ClfFcn$, which is trained with standard cross entropy on labeled samples $\mat{X}_L$, retaining its efficient \emph{inductive} nature.

\subsection{Encouraging Compact Clusters in Feature Space}
\label{subsec:loss_func}

\begin{algorithm}[!ht]
   	\caption{Training for SSL with CCLP}
   	\label{alg:algo1}
\begin{algorithmic}
   	\STATE {\bfseries Input:} feature extractor $\FeatFcn(\cdot; \FeatParam)$, classifier $\ClfFcn(\cdot; \ClfParam)$, \\
   	data $\Dataset_L$, $\Dataset_U$, batch sizes $N_L$, $N_U$, $N \!=\! N_L \!+\! N_U$\\
   	Markov chain steps $S$, weighting $w$, learning rate $\alpha$ \\
   	\STATE {\bfseries Output:} Learnt network parameters $\FeatParam$ and $\ClfParam$
   	\REPEAT
   		\STATE $(\mat{X}_L, \vec{y}_L)\! \overset{N_L}{\sim}\! \Dataset_L,\ \mat{X}_U \! \overset{N_U}{\sim}\! \Dataset_U, \ \mat{X}\!=\! \begin{bsmallmatrix} \mat{X}_L \\ \mat{X}_U \end{bsmallmatrix}$ \# Samples
   		\STATE $\mat{Y}_L \leftarrow \mathtt{one\_hot}( \vec{y}_L )$ \quad\# Labels
   		\STATE $\mat{Z} \leftarrow \FeatFcn(\mat{X}; \FeatParam)$ \quad\# Forward pass
   		\STATE $\Loss{sup} \leftarrow - \frac{1}{N_L} \sum_{i=1}^{N_L} \sum_{c=1}^C y_{ic} \log [\ClfFcn( \vec{z}_i; \ClfParam )]_c$
   		\vspace{5px}
   		\STATE $\mat{A} \leftarrow \exp( \mat{Z} \mat{Z}\tr )$ \quad\# Graph
   		\STATE $\mat{H} \leftarrow $ row normalized $\mat{A}$ \quad\# Transition matrix
   		\STATE $\bPhi_U \leftarrow (\mat{I} - \mat{H}_{UU})^{-1} \mat{H}_{UL} \mat{Y}_L$ \quad\# LP
   		\STATE $\bPhi \leftarrow [ \mat{Y}_L; \bPhi_U ]$
   		\STATE $\mat{T} \leftarrow $ according to \cref{eq:target}
   		\STATE $\mat{M} \leftarrow \bPhi \bPhi\tr$ \quad\# Labels agreement
   		\STATE $\Loss{CCLP} \leftarrow 0$
   		\FOR{$s=1$ {\bfseries to} $S$}
   			\STATE $\mat{H}^{(s)} \leftarrow (\mat{H} \circ \mat{M})^{s-1} \mat{H}$
			\STATE $\Loss{CCLP} \leftarrow \Loss{CCLP} - \frac{1}{S N^2} \sum_{i=1}^{N} \sum_{j=1}^{N} T_{ij} \log H_{ij}^{(s)}$
		\ENDFOR
   		\STATE $\Loss{total} \leftarrow \Loss{sup} + w\Loss{CCLP}$
   		\STATE $\FeatParam \leftarrow \FeatParam - \alpha \frac{\partial \Loss{total}}{\partial \FeatParam}$, $\quad$ $\ClfParam \leftarrow \ClfParam - \alpha \frac{\partial \Loss{sup}}{\partial \ClfParam}$ \# Updates
	\UNTIL{stopping criterion is $true$}\\
\end{algorithmic}
\end{algorithm}

Our desiderata for an optimal SSL regularizer are as follows: 1) it encourages formation of a single and compact cluster per class in latent space, so that linear separation is straightforward; 2) it must be compatible with the supervised loss to allow easy optimization and high performance.

We first observe that in the desired optimal state, where a single, compact cluster per class has been formed, the transition probabilities between any two samples of the same class should be the same, and zero for inter-class transitions. Motivated by this, we define a soft version of this optimal transition matrix $\mat{T}$ as:
\begin{equation}\label{eq:target}
T_{ij} = \sum_{c=1}^{C} \phi_{ic} \frac{ \phi_{jc}}{ m_c } \,, \quad m_c = \sum_{i=1}^{N} \phi_{ic} \,.
\end{equation}
Here $\phi_{ic}$ is the LP posterior for node $i$ to belong to class $c$ and $m_c$ the expected mass assigned to class $c$. We encourage $\FeatFcn$ to form compact clusters by minimizing cross entropy between the ideal $\mat{T}$ and the current transition matrix $\mat{H}$:
\begin{equation}\label{eq:loss_diff_1step}
\Loss{1\!-\!step} = \frac{1}{N^2} \sum_{i=1}^{N} \sum_{j=1}^{N} -T_{ij} \log H_{ij} \,.
\end{equation}

This objective has properties that are desirable in SSL. It considers unlabeled samples, models high and low density regions via the use of LP, and facilitates separation by attracting together in one compact cluster samples of the same soft or hard labels while repulsing different ones. It does not apply strong forces to unconfident samples, to avoid problematic optimization when embedding is still suboptimal, e.g.~in early training. By being unaware of $\ClfFcn$ and its decision boundary, gradients of \cref{eq:loss_diff_1step} only depend on the manifold's geometry, thus they do not oppose those from the supervised loss, unlike methods suffering from confirmation bias (\cref{sec:related}). We argue that one more property is important for good optimization, which is not yet covered.

During optimization, forces applied by a cost should not disturb existing clusters, as they contain information that enables SSL via the cluster assumption. To model such behavior we design a cost that attracts points of the same class along the structure of the graph. For this we extend the regularizer of \cref{eq:loss_diff_1step} to the case of Markov chains with multiple transitions between samples, which should remain within a single class. The probability of a Markov process with transition matrix $\mat{H}$ starting at node $i$ and landing at node $j$ after $s$ number of steps is given by $\left( \mat{H}^s \right)_{ij}$.

We are interested in modeling transitions within the same class and increase their probability, while minimizing the probability of transiting to other clusters. Our solution is to utilize the class posteriors estimated by LP, to define a confidence metric that nodes belong to the same class. For this, we use the dot product of the nodes' LP class posteriors $\mat{M} = \bPhi \bPhi\tr$. The convenient property of this choice is that the elements of $\mat{M}$ are bounded in the range $[0,1]$, taking the maximum and minimum values if and only if the labels (hard/soft for $\mat{X}_L$/$\mat{X}_U$ respectively) fully agree or disagree respectively. This allows us to use it as an estimate of the probability that two nodes belong to the same class.

Equipped with $\mat{M}$, we estimate the joint probability of transitioning from node $i$ to node $j$ and the two belonging to the same class as ${p(i \rightarrow j, y_i=y_j)}={p(i \rightarrow j)} {p(y_i = y_j | i \rightarrow j)} \approx H_{ij}M_{ij} $. Note that $\mat{M}$ is indeed a function of $\mat{H}$, as suggested by the conditional it estimates. Finally, we estimate the probability of a Markov process to start from node $i$, perform $(s\!-\!1)$ steps within the same class and then transit to \emph{any} node $j$, as the element $H^{(s)}_{ij}$ of the matrix
\begin{equation}\label{eq:multistep_cluster_H}
\mat{H}^{(s)} = (\mat{H} \circ \mat{M})^{s-1} \mat{H} = (\mat{H} \circ \mat{M}) \mat{H}^{(s-1)},
\end{equation}
where $\circ$ denotes the Hadamard (elementwise) product.

By regularizing $\mat{H}^{(s)}$ towards target matrix $\mat{T}$ as in~\cref{eq:loss_diff_1step}, we minimize the probability of a chain transiting between clusters of different classes after $s$ steps, thus repulsing them, and encourage uniform probability for chains that only traverse samples of one class, which attracts them and promotes compact clustering. Notably, regularizing $\mat{H}^{(s)}$ of the latter type of chains towards larger values discourages disturbing clusters along their path, as this would push $\mat{H}^{(s)}$ close to zero. This motivates the final form of our \textit{Compact Clustering via Label Propagation} (CCLP) regularizer:
\begin{equation} \label{eq:loss_cclp}
\Loss{CCLP} = \frac{1}{S} \sum_{s=1}^S
			\frac{1}{N^2} \sum_{i=1}^{N} \sum_{j=1}^{N} - T_{ij} \log H^{(s)}_{ij}.
\end{equation}
This cost consists of $S$ terms, each modeling paths of different length between samples on the graph. Larger $s$ allows Markov chains to traverse along more elongated clusters. Note that this cost subsumes \cref{eq:loss_diff_1step} for $s=1$. \Cref{eq:loss_cclp} is minimized simultaneously with the supervised loss $\Loss{sup}$. An overview of our method is shown in Algorithm~\ref{alg:algo1}.



\section{Empirical Analysis on Synthetic Data}
\label{sec:analysis}

\begin{figure*}[!ht]
\centering
\includegraphics[width=\linewidth]{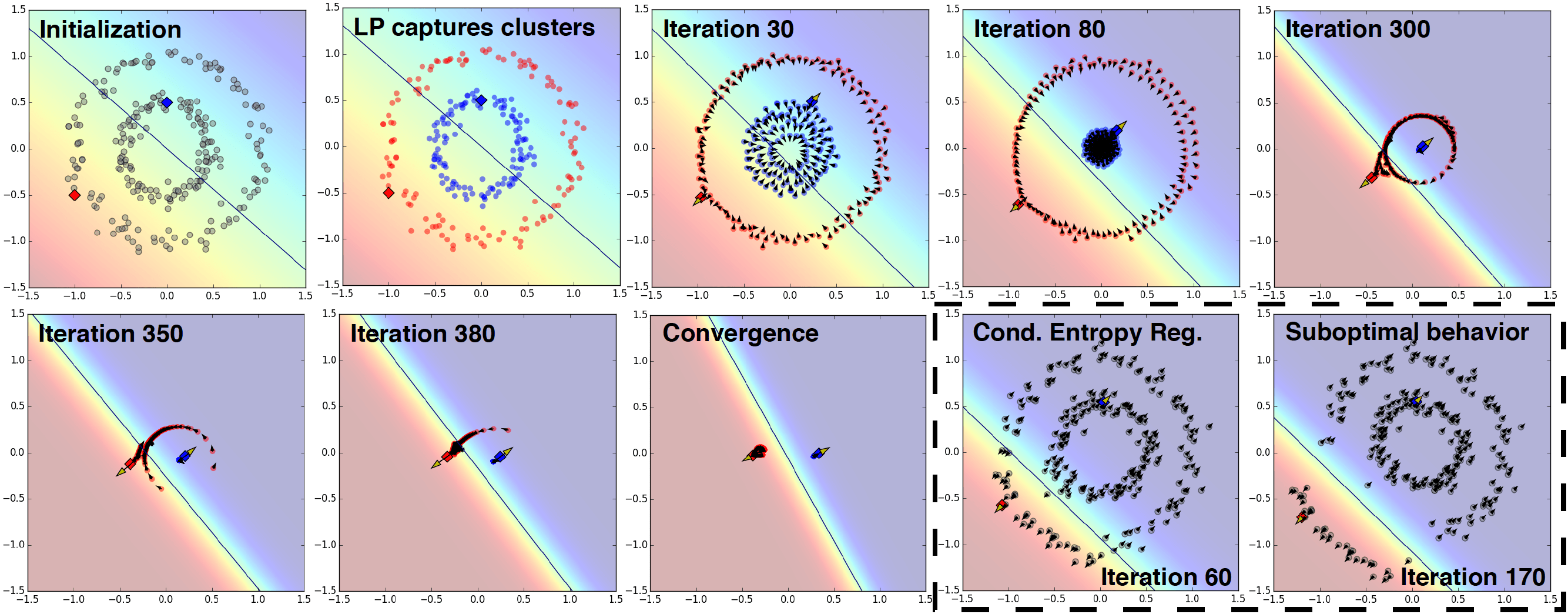}
\vspace{-5mm}
\caption{Two-circles toy experiment. Main figure shows the initial arrangement of two labeled (red/blue) and multiple unlabeled points, label propagation, and iterations using CCLP along with supervision until convergence. Also depicted are gradients by the supervised loss on labeled samples (yellow arrows) and by CCLP (black arrows). The dashed box shows failure case of conditional entropy regularizer.}
\label{fig:two_circles}
\end{figure*}

We conduct a study on synthetic toy examples to analyze the behavior of the proposed method. We are interested in the forces that CCLP applies to samples in the latent space $\FeatSpace$ as it attempts to improve their clustering, \emph{isolated} from the influence of model $\FeatFcn(\cdot;\FeatParam)$. Hence we do not adopt common visualization methods that map space $\FeatSpace$ learned by a network to 2D \cite{maaten2008visualizing}, or plot the decision boundary of the total model $f$ in input space $\mathcal{X}$.

To isolate the effect of CCLP, we consider an artificial setup in which assumed embeddings of samples $\mat{Z}$ are initially positioned in a structured arrangement in a 2D space, which represents $\FeatSpace$, and are allowed to move freely. We place the embeddings in commonly used toy layouts: two-moons and two-circles. For the role of $\ClfFcn( \mat{Z}; \ClfParam)$, we use a linear classifier, for which we compute the supervised loss $\Loss{sup}$. We then perform label propagation on this artificial latent space $\FeatSpace$ and compute $\Loss{CCLP}$. Finally, we compute the gradients of the two costs with respect to $\ClfParam$ and the coordinates of the embeddings $\mat{Z}$, and update them iteratively.

In this setting, both costs try to move the labeled samples in space $\FeatSpace$, but only CCLP affects unlabeled data. If $\mat{Z}$ were computed by a real neural net $\FeatFcn(\cdot; \FeatParam)$, which is a smooth function, embeddings of unlabeled samples would also be affected by $\Loss{sup}$, via updates to $\FeatParam$. Our settings instead isolate the effect of CCLP on the unlabeled data. 

\subsection{Two Circles}

We study the dynamics of CCLP ($S \!=\! 10$) on two-circles (\cref{fig:two_circles}), when a single labeled example is given per class. We first observe that the isolated effect of CCLP indeed encourages formation of a single, compact cluster per class. In more challenging scenarios, results are naturally subject to the effect of the model, optimizer and data.

We also observe that the direction of gradients applied by CCLP to each sample depends on the manifold's geometry, not on the decision boundary, about which CCLP is agnostic. Since gradients from the supervised loss are perpendicular to the decision boundary of the linear classifier, the effect of CCLP generally does not oppose supervision. By contrast, we show the effect of confirmation bias by studying conditional entropy regularization (CER) \cite{grandvalet2005semi}. CER gradients are perpendicular to the decision boundary and can thus oppose the effect of supervision.\footnote{If combined with an appropriate model $\FeatFcn$ or a different optimizer, CER could solve this example. Here we focus on the effect under gradient descent and independently of the model $\FeatFcn$.}

\subsection{Two Moons}

\begin{figure*}[!ht]
\centering
\includegraphics[width=\linewidth]{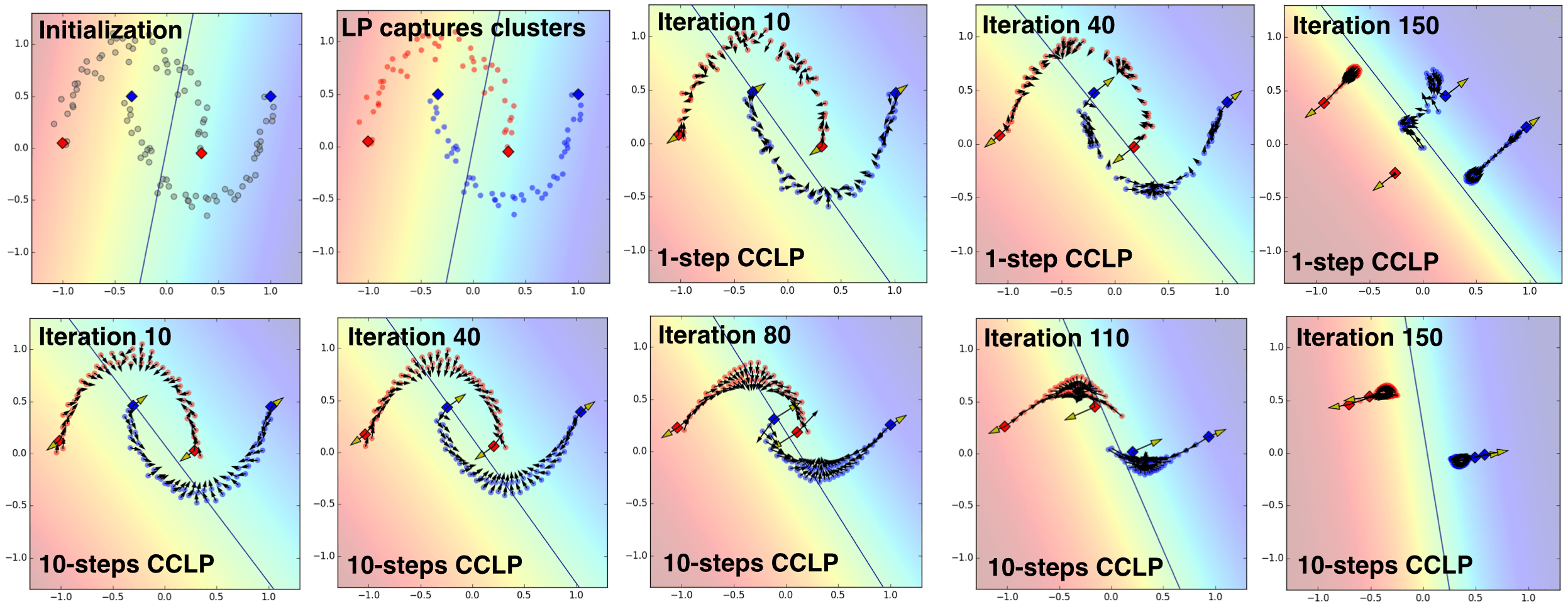}
\vspace{-5mm}
\caption{Two-moons toy experiment. Comparison between CCLP applied with $S \!=\! 1$ (\emph{top row}) and with $S \!=\! 10$ (\emph{bottom row)}. Exploring the direction of the gradients from CCLP (black arrows) shows that optimizing over a longer chain of steps leads to a behavior that tries to preserve existing clusters when attempting to create more compact clusters.}
\label{fig:two_moons}
\end{figure*}

We use two moons to investigate the effect of the maximum steps $S$ of Markov chains used in $\Loss{CCLP}$ (Fig.~\ref{fig:two_moons}). When multiple steps are used, here $S \!=\! 10$, gradients of CCLP follow existing high density areas in their attempt to cluster samples better. This leads the labeled samples to also move along the existing clusters on their way to the correct side of the decision boundary. Conversely, when a single step is used ($S \!=\! 1$), gradients by CCLP try to preserve only local neighborhoods, which allows the clusters to disintegrate. This breakdown of the global structure implies loss of information, which in turn may lead to misclassification.

\section{Evaluation on Common Benchmarks}
\label{sec:main_eval}

\begin{table*}[!ht]
\sisetup{
	table-number-alignment=center,
	separate-uncertainty=true,
	table-figures-uncertainty=2,
	table-figures-integer=1,
	table-figures-decimal=2,
    mode=text
}

\fontsize{5.5}{5.5}\selectfont
\caption{Performance of CCLP compared to contemporary SSL methods on common benchmarks, when limited or all available labelled data is used as $\Dataset_L$ for training. Also shown is performance of the corresponding baseline with standard supervision (no SS). Error rate is shown as (mean $\pm$ st.dev.). Only results obtained without augmentation are shown. Methods in the lower part used larger classifiers.}
\label{tab:main_results}
\begin{center}
\begin{sc}
\renewcommand{\arraystretch}{1.1}
\begin{tabular}{lSSSSSSSSS}
\toprule
											& \multicolumn{3}{c}{MNIST}						& \multicolumn{3}{c}{SVHN}						& \multicolumn{3}{c}{CIFAR10} \\
											  \cmidrule(lr){2-4}							  \cmidrule(lr){5-7}							  \cmidrule(lr){8-10} 
Model										& {$|\Dataset_L|=100$} & {All}	& {All, No SS}	& {$1000$}		& {All}			& {All, No SS}	& {$4000$}		& {All}			& {All, No SS} \\
\midrule
conv-CatGAN\cite{springenberg2015} 			& 1.39 \pm 0.28	& 0.48 			& {--}			& {--}			& {--}			& {--}			& 19.58\pm 0.46	& 9.38 			& {--} \\
Ladder(CNN-$\Gamma$)\cite{rasmus2015semi}	& 0.89 \pm 0.50	& {--} 			& 0.36			& {--}			& {--}			& {--}			& 20.40\pm 0.47	& {--} 			& 9.27 \\
SDGM \cite{maaloe2016auxiliary} 			& 1.32 \pm 0.07	& {--} 			& {--} 			& 16.61\pm 0.24	& {--}			& {--}			& {--}			& {--} 			& {--} \\
ADGM \cite{maaloe2016auxiliary}				& 0.96 \pm 0.02	& {--} 			& {--}			& 22.86			& {--}			& {--}			& {--}			& {--} 			& {--} \\
iGAN \cite{salimans2016improved}			& 0.93 \pm 0.07	& {--}			& {--}			& 8.11 \pm 1.3	& {--}			& {--}			& 18.63\pm 2.32	& {--} 			& {--} \\
ALI	\cite{dumoulin2016adversarially}		& {--}			& {--} 			& {--}			& 7.42 \pm 0.65	& {--} 			& {--}			& 17.99\pm 1.62	& {--} 			& {--} \\
VAT	 \cite{miyato2017virtual}				& 1.36			& 0.64 			& 1.11			& 6.83			& {--}			& {--}			& 14.87			& 5.81 			& 6.76 \\
triple GAN \cite{chongxuan2017triple}		& 0.91 \pm 0.58	& {--} 			& {--}			& 5.77 \pm 0.17	& {--}			& {--}			& 16.99\pm 0.36	& {--} 			& {--} \\
mmCVAE \cite{li2017max}						& 1.24 \pm 0.54	& 0.31 			& {--}			& 4.95 \pm 0.18	& 3.09			& {--}			& {--}			& {--} 			& {--} \\
BadGan \cite{dai2017good}					& 0.80 \pm 0.10	& {--} 			& {--}			& 4.25 \pm 0.03	& {--}			& {--}			& 14.41\pm 0.30	& {--} 			& {--} \\
LBA	\cite{haeusser2017learning}				& 0.89 \pm 0.08	& 0.36 \pm 0.03 & {--}			& {--}			& {--}			& {--}			& {--} 			& {--}			& {--} \\
LBA (our implementation)					& 0.90 \pm 0.10	& 0.36 \pm 0.03	& 0.46 \pm 0.03	& 9.25 \pm 0.65 & 3.61 \pm 0.10 & 4.26 \pm 0.10	& 19.33\pm 0.51 & 8.46 \pm 0.18	& 9.33 \pm 0.14 \\
\textbf{CCLP (ours)}						& 0.75 \pm 0.14	& 0.32 \pm 0.03	& 0.46 \pm 0.03	& 5.69 \pm 0.28	& 3.04 \pm 0.05 & 4.26 \pm 0.10	& 18.57\pm 0.41	& 8.04 \pm 0.18	& 9.33 \pm 0.14 \\
\midrule
\emph{Larger Classifiers}					&				&				&				&				& 				&				& 				& 				& \\
$\Pi$ model	\cite{laine2016temporal}		& {--}			& {--}			& {--}			& 5.43 \pm 0.25	& {--} 			& {--}			& 16.55\pm 0.29	& {--}			& {--} \\
MTeach.\cite{tarvainen2017mean}				& {--}			& {--}			& {--}			& 5.21 \pm 0.21	& 2.77 \pm 0.09 & 3.04 \pm 0.04	& 17.74\pm 0.29	& 7.21 \pm 0.24	& 7.43 \pm 0.06 \\
VAT-large \cite{miyato2017virtual}			& {--}			& {--} 			& {--}			& 5.77			& {--}			& {--}			& 14.82			& {--}			& {--} \\
VAT-large-Ent \cite{miyato2017virtual}		& {--}			& {--} 			& {--}			& 4.28			& {--}			& {--}			& 13.15			& {--}			& {--} \\

\bottomrule
\end{tabular}
\end{sc}
\end{center}
\vskip -0.1in
\end{table*}

\textbf{Benchmarks:} We consider three benchmarks widely used in studies on SSL: MNIST, SVHN and CIFAR-10. Following common practice we whiten the datasets. We use no data augmentation to isolate the effect of the SSL method. Following previous work, to study the effectiveness of CCLP when labeled data is scarce, as $\Dataset_L$ we use 100, 1000 and 4000 samples from the training set of each benchmark respectively, while the whole training set without its labels constitutes $\Dataset_U$. We also study effectiveness of our method when abundant labels are available, using the whole training set as both $\Dataset_L$ and $\Dataset_U$. We also report performance of our baseline trained with only standard supervision (no SSL), to facilitate comparison of improvements from CCLP with previous and future works, where quality of the baselines may differ. For every benchmark, we perform $10$ training sessions with random seeds and randomly sampled $\Dataset_L$ and report the mean and standard deviation of the error. We evaluate on the test-dataset of each benchmark, except for the ablation study where we separated a validation set.

\textbf{Models:} For MNIST we use a CNN similar to \citet{rasmus2015semi,chongxuan2017triple,haeusser2017learning,li2017max}. For SVHN and CIFAR we use the network used as classifier in \citet{salimans2016improved}, commonly adopted in recent works. In all experiments we used the same meta-parameters for CCLP: In each SGD iteration we sample a batch $(\mat{X}_L, \vec{y}_L) \!\sim\! \Dataset_L $ of size $N_L \!=\! 100$, where we ensure that 10 samples from each class are contained, and a batch without labels $\mat{X}_U \!\sim\! \Dataset_U$ of size $N_U \!=\! 100$. We use the dot product as similarity metric (\cref{eq:adjacency_mat}), $S \!=\! 3$ maximum steps of the Markov chains (\cref{eq:loss_cclp}). $\Loss{CCLP}$ was weighted equally with the supervised loss, with $w \!=\! 1$ throughout training. These parameters were found to work reasonably in early experiments on a pre-selected validation subset from the MNIST training set and were used without extensive effort to optimize them for this benchmarking. Exception are the experiments with $|\Dataset_L| \!=\! 4000$ on CIFAR, where lower $w \!=\! 0.1$ was used, because $w \!=\! 1$ was found to over-regularize these settings.

We also employ the method of \citet{haeusser2017learning} (LBA) on SVHN and CIFAR-10, as in the original work a different network was used, while results on SVHN where reported only with data augmentation. We note that for correctness, in preliminary experiments we ensured that with data augmentation our implementation of LBA produced similar results to what reported in \citet{haeusser2017learning}.

\textbf{Results:} Performance of our method in comparison to recent SSL approaches that use similar experimental settings are reported in \cref{tab:main_results}. We do not report results obtained with data augmentation. Note that iGAN \cite{salimans2016improved}, VAT \cite{miyato2017virtual} and BadGAN \cite{dai2017good} used a deep MLP instead of a CNN on MNIST, so those results may not be entirely comparable. Our method achieves very promising results in all benchmarks, that improve or are comparable to the state-of-the-art. CCLP consistently improves performance over standard supervision even when all labels in the training set are used for supervision, indicating that CCLP could be used as a latent space regularizer in fully supervised systems. In the latter settings, CCLP offers greater improvement over the corresponding baselines than the most recent perturbation-based method, mean teacher \cite{tarvainen2017mean}. We finally emphasize that our method consists of the computation of a single cost function and does not require additional network components, such as the generators required for VAEs and GANs, or the density estimator PixelCNN++ used in \citet{dai2017good}. Furthermore, we emphasize that many of these works make use of multiple complementary regularization costs. The compact clustering that our method encourages is orthogonal to previous approaches and could thus boost their performance further.

\begin{figure}[!ht]
\centering
\includegraphics[width=\linewidth]{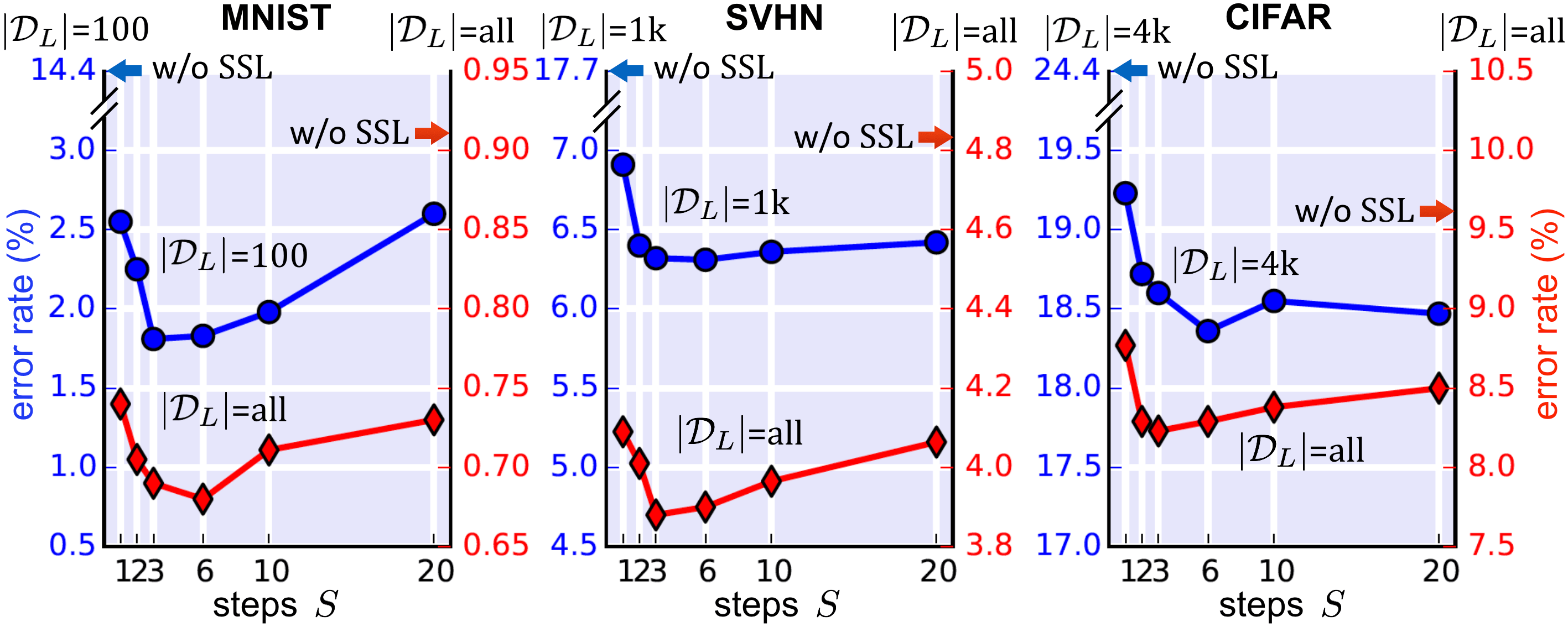}
\vspace{-5mm}
\caption{Validation error when CCLP is applied with varying number of steps $S$. Left/right vertical axis correspond to training with limited/all labeled samples respectively. Compact clustering with $S\!=\!1$ improves over standard supervision (w/o SSL). Optimizing with 3-6 steps offers a further $5$-$25\%$ reduction of the error.}
\label{fig:ablation}
\end{figure}

\textbf{Ablation study:} We further study the effect of CCLP's two key aspects: Regularizing the latent space towards compact clustering and, secondly, optimizing while respecting existing clusters by using multi-step Markov chains. For this, we separate a validation set of 10000 images from the training set of each benchmark. $\Dataset_L$ and $\Dataset_U$ are formed out of the remaining training data. We evaluate performance on the validation set when CCLP uses different number of maximum steps $S$ (\cref{eq:loss_cclp}). Each setting is repeated 10 times and we report the average error in Fig.~\ref{fig:ablation}. When $S\!=\!1$, CCLP encourages compact clustering without attempting to preserve existing clusters. This already offers large benefits over standard supervision. Optimizing over longer Markov chains offers further improvements, with values $3\! \leq \! S \! \leq \! 6$ further reducing the error by $5$-$25\%$ in most settings. Capturing too long paths between samples ($S\!>\!10$) reduces the benefits.


\section{Computational Considerations}

Time complexity of CCLP is $O(N^3 \!+\! S N^2)$, overwhelmed by $O(N^3)$ of matrix inversion since $N\! \gg \!S$ in our settings. In practice, CCLP is inexpensive compared to a net's forward and backward passes. In our CIFAR settings and \mbox{TensorFlow} GPU implementation \cite{abadi2016tensorflow}, CCLP increases less than $10\%$ the time for an SGD iteration, even for large $N\!=\!1000$. In comparison, GANs and VAEs require an expensive decoder, while perturbation-based approaches perform multiple passes over each sample.

As batch size $N$ defines how well the graph approximates the true data manifold, larger $N$ is desirable but requires more memory, while low $N$ may decrease performance. Batch sizes in the order of $200$ used in this and previous works \cite{laine2016temporal} are practical in various applications, with hardware advances promising further improvements. Finally, in distributed systems that divide thousands of samples between compute nodes, CCLP could scale by creating a different graph per node.


\section{Conclusion}

We have presented a novel regularization technique for SSL, based on the idea of forming compact clusters in the latent space of a neural network while preserving existing clusters during optimization. This is enabled by dynamically constructing a graph in latent space at each SGD iteration and propagating labels to estimate the manifold's structure, which we then regularize. We showed that our approach is effective in leveraging unlabeled samples via empirical evaluation on three widely used image classification benchmarks. We also showed our regularizer offers consistent improvements over standard supervision even when labels are abundant.

Our method is computationally efficient and easy to apply to existing architectures as it does not require additional network components. It is also orthogonal to approaches that do not capture the structure of data, such as perturbation based approaches and self-supervision, with which it can be readily combined.
Analyzing further the properties of a compactly clustered latent space, as well as applying our approach to larger benchmarks and the task of semantic segmentation is interesting future work.


\section*{Acknowledgements}

This research was partly carried out when KK, DC and RT were interns at Microsoft Research Cambridge. This project has also received funding from the European Research Council (ERC) under the European Union's Horizon 2020 research and innovation programme (grant agreement No 757173, project MIRA, ERC-2017-STG). KK is also supported by the President's PhD Scholarship of Imperial College London. DC is supported by CAPES, Ministry of Education, Brazil (BEX 1500/15-05).
LLF is funded through EPSRC Healthcare Impact Partnerships grant (EP/P023509/1). IW is supported by the Natural Environment Research Council (NERC).

\bibliography{bibliography}
\bibliographystyle{icml2018}



\end{document}